# Deep Learning architectures for generalized immunofluorescence based nuclear image segmentation

Florian Kromp, Lukas Fischer, Eva Bozsaky, Inge Ambros, Wolfgang Doerr, Sabine Taschner-Mandl, Peter Ambros, Allan Hanbury

*Abstract*—Separating and labeling each instance of a nucleus (instance-aware segmentation) is the key challenge in segmenting single cell nuclei on fluorescence microscopy images. Deep Neural Networks can learn the implicit transformation of a nuclear image into a probability map indicating the class membership of each pixel (nucleus or background), but the use of post-processing steps to turn the probability map into a labeled object mask is error-prone. This especially accounts for nuclear images of tissue sections and nuclear images across varying tissue preparations. In this work, we aim to evaluate the performance of state-of-the-art deep learning architectures to segment nuclei in fluorescence images of various tissue origins and sample preparation types without post-processing. We compare architectures that operate on pixel to pixel translation and an architecture that operates on object detection and subsequent locally applied segmentation. In addition, we propose a novel strategy to create artificial images to extend the training set. We evaluate the influence of ground truth annotation quality, image scale and segmentation complexity on segmentation performance. Results show that three out of four deep learning architectures (U-Net, U-Net with ResNet34 backbone, Mask R-CNN) can segment fluorescent nuclear images on most of the sample preparation types and tissue origins with satisfactory segmentation performance. Mask R-CNN, an architecture designed to address instance aware segmentation tasks, outperforms other architectures. Equal nuclear mean size, consistent nuclear annotations and the use of artificially generated images result in overall acceptable precision and recall across different tissues and sample preparation types.

*Index Terms*—nuclei segmentation, artificial datasets, deep learning, architecture comparison

## Introduction

**M**ICROSCOPY has become a powerful tool to gain insights into cellular or sub-cellular structures by visualizing cellular compartments such as the nucleus, the cytoplasm, sub-cellular appearance of proteins or DNA elements [1]. By using automated microscopes and by applying image analysis workflows, quantitative results can be generated at the single cell level. These allow the detection of even subtle biological changes while taking advantage of the statistical power of analyzing thousands of cells. The main sites of operation for quantitative microscopy analysis are pathology departments and diagnostic laboratories. In addition, quantitative microscopy techniques are applied and refined in research laboratories. While pathology departments routinely use Hematoxylin and eosin (H&E) histological or immuno-histochemical (IHC) stainings, research laboratories mainly rely on immunofluorescence (IF) stainings. This is because up to 90 or more (sub-)cellular compartments can be visualized simultaneously using multiplex-IF staining techniques and epifluorescence microscopy. This provides a substantial gain of information compared to the visualization of two to three cellular characteristics when using H&E or IHC stainings and brightfield microscopy. While pathology departments mainly rely on tissue sections to diagnose disease type and grade or stage of cancer [2], [3], research laboratories frequently use additional tissue preparations such as cell lines grown on or cytospinned to microscopy glass slides, bone marrow cytospins or tumor touch imprints. Regardless of the tissue origin or type of sample preparation, quantitative, microscopy based image analysis workflows generally consist of the following steps: sample preparation, microscopy image acquisition, nuclear and/or cytoplasmic image segmentation, feature extraction and cell population analysis. Each step within such a workflow can have a significant impact on quantification and thus, on interpretation of experiments [4].

A prerequisite that also depicts a bottleneck in automated quantitative microscopy is to obtain a satisfactory nuclear image segmentation accuracy. To generate quantitative analysis results at the single cell level, segmentation algorithms must segment each nucleus instance. Such algorithms are frequently called *instance segmentation* or *instance aware segmentation* algorithms. Inaccurate image segmentation, frequently caused by tightly aggregated nuclei that cannot be separated by the segmentation algorithm, can lead to deriving wrong conclusions from experiments [5]. To solve the task of separating nuclear instances in aggregations, post-processing steps are usually applied to an initial nucleus segmentation. Two main approaches are commonly used: image over-segmentation with subsequently applied region merging or image under-segmentation with subsequently applied splitting of aggregations. Usually, post-processing steps rely on morphological or intensity based features derived from segmented objects. These features are used to identify over- and under-segmented objects and to guide the separation or merging process.

Still, nuclear segmentation algorithms frequently fail to segment nuclear instances due to the following reasons:

- Epifluorescent microscopic images are blurry caused by out-of-focus light, leading to aggregations if nuclei are located in spatially close neighborhoods.
- Nuclei, in dense tissue sections are frequently aggregated and/or overlapped and present arbitrary convex or even concave shapes due to 2D representations of 3D specimens, aggravating the use of morphological based features for applying post-processing operations.

- Bone marrow cytospins and tumor touch imprints show aggregated nuclei of different cell types that occasionally show heterogeneus nuclear intensities, preventing the use of intensity based features for applying post-processing operations.

Figure 1 shows an example of images highly challenging human experts as well as automated nuclear image segmentation methods on separating each nucleus instance.

To the best of our knowledge, there is no algorithm published that solves an instance aware segmentation of nuclei on fluorescent images of various tissue origins, sample preparation types and magnifications, further called sample conditions. This might be based on two reasons: 1. Post-processing operations are usually based on parameters that are optimized to a certain sample condition. This hampers the overall aim to obtain a satisfactory segmentation performance on nuclear images across varying sample conditions. 2. Expert-annotated fluorescent nuclear image datasets that cover a broad range of sample conditions and that can be used to train machine-learning based algorithms were not available.

Hence, we recently published an expert-annotated dataset consisting of fluorescent nuclear images and annotations of various tissue origins and sample preparation types, acquired using different levels of magnification [6]. We hypothesize that by training deep learning based image segmentation architectures using this dataset, they can learn to accurately segment nuclear instances in challenging images across varying sample conditions.

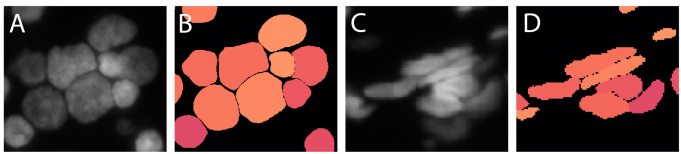

Fig. 1. Examples of nuclear morphologies in different tissue preparations. A. Neuroblastoma bone marrow cytospin presenting varying nuclear intensity and size. B. Annotated mask of A. C. Ganglioneuroma tissue cryosection presenting overlapping/aggregated nuclei with varying morphology and intensity. D. Annotated mask of C.

### A. Contributions

We developed and applied a pipeline to compare the segmentation performance of multiple deep learning architectures trained using the aforementioned expert-annotated dataset. In combination with a novel strategy developed to extend the dataset by artificially generating images, we demonstrate that nuclear images of different tissue origins and sample preparation types can be segmented with a satisfactory segmentation performance, without the need to apply specific post-processing operations that are optimized to fit a specific sample condition. Moreover, we provide evidence that multiple factors influence the segmentation performance including the quality of dataset annotations, image scales, image segmentation challenge levels, sample preparation types and the framework used.

## RELATED WORK

We briefly describe deep learning based segmentation approaches targeting brightfield nuclear images. We outline the difference to fluorescence images and describe deep learning based approaches operating on fluorescence nuclear images. Moreover, we describe recently published data augmentation strategies and methods generating synthetic images to extend nuclear image datasets.

The most popular nuclear segmentation algorithms designed until deep learning (DL) architectures gained importance are based on the watershed algorithm, region growing, level-set or active contour methods (for a comprehensive overview see [7]).

### Deep learning based methods to segment nuclear images of H&E and IHC stained samples

Recent work showed that Deep Convolutional Neural Networks (DCNN) outperformed most standard methods applied in computer vision tasks such as image classification or segmentation by a large margin [8], [9]. Feature representations are learned by supervised training such that they do not have to be designed and tuned manually to fit a given problem. DCNNs have to be trained using large datasets to adjust the parameters of each network layer targeting the minimization of a loss function. In biomedical sciences, large annotated datasets are usually not available, and thus networks were designed to enable learning from small datasets. Ronneberger et al. proposed a network structure called U-Net focusing on an application to biomedical image segmentation problems providing only small datasets [10]. Cui et al. [11] use a modified U-Net structure to segment H&E slides of challenging tissue sections. Recently, annotated datasets of H&E and IHC stained samples are publicly available. This led to the development of new deep learning based architectures to segment challenging nuclear images of H&E or IHC stained tissue sections [11], [12], [13]. Naylor et al. [14] use and compare CNN architectures (FCN, U-Net, Mask R-CNN) for segmenting H&E stained histological slides. They formulate the segmentation problem as a regression task, leading to a prediction of the distance-transform of the binarized image.

### Deep learning based methods to segment immunofluorescence based nuclear images

In contrast to annotated H&E or IHC stained nuclear image datasets, only a limited number of annotated fluorescence nuclear images can be obtained publicly. Moreover, they do not cover images of tissue sections or varying sample preparation types. As deep learning based segmentation methods rely on annotated datasets, there are no methods published that build upon deep learning approaches to segment nuclear images of immunofluorescent stained tissue specimens. To the best of our knowledge there is only one publication available presenting a deep learning based approach to segment nuclear images of samples of arbitrary tissue origin and staining types. Alom et al. [13] use a Recurrent Residual Convolutional Neural Network based U-Net to segment images of the 2018 Data

Science Bowl Grand Challenge dataset, including, among others, nuclear images of H&E stained tissue sections and of immunofluorescently stained cells grown on microscopy glass slides. However, results indicate that the proposed architecture does not ensure instance aware segmentation.

Other deep learning based segmentation methods operate on datasets showing lower segmentation complexity when compared to the segmentation of tissue sections, indicated by a low number of nuclear aggregations or overlaps, involving only a few cells, or operate on 3D image stacks. Caicedo et al. [15] compare a U-Net and the DeepCell architecture [16] evaluated on images of the BBBC022 dataset as part of the Broad Bioimage Benchmark Collection, a dataset consisting of nuclear images of cells grown on microscopy slides. Ho et al. [17] use a SegNet architecture [18] to segment 3D image stacks of rat kidney tissue. Images of lung carcinoma cells grown on slides and images from the BBBC022 dataset were used to evaluate a FCN network structure by Sadanandan et al. [19].

### Data augmentation and artificial image synthesis

Data augmentation techniques can substantially improve the prediction performance of deep neural networks [20]. On biomedical image segmentation tasks, data augmentation was introduced by Ronneberger et al. [10] in 2014. Recently, Cui et al. showed the benefit of data augmentation in nuclear segmentation of H&E stained histopathological samples [11]. Despite the value of data augmentation techniques applied to deep learning training sets, IF based imaging is subject to multiple parameters not addressed by data augmentation techniques. Parameters include varying image integration time and varying quality and intensity of a given immuno-staining signal. Weak signals have to be captured with higher integration time to ensure an acceptable dynamic range of the resulting images, leading to an overall increased background intensity and thus, a low signal-to-noise ratio. Moreover, the intensity of nuclei can vary within an observed field of view (FOV), depending on the amount of DNA, the intactness of a cell and whether a nucleus is in focus or not. These variation in the appearance of nuclei is not reflected by standard data augmentation techniques that purely apply transformations (flipping, cropping, rotation, elastic deformations, intensity variations, shifting and more) on training set images.

To allow for a better generalization performance in fluorescence image nuclei segmentation, the use of synthetic datasets was proposed. Russell et al. [21] created simulated images by modeling nuclear shape and fluorescence image characteristics by overlaying the image with Gaussian noise and blurring the image. Hou et al. [22] proposed a pipeline using real image patches from histo-pathological images to create artificial image patches. First, they segmented all nuclei of an image patch and removed them from the patch to create a nucleus-free patch. Then, they created artificial nuclei utilizing randomly shaped polygons, sampled from a predefined distribution, and placed them on the nucleus-free patch. Finally, they used a neural network called *refiner CNN* trying to match the image with a reference style obtained from original nuclear images. By segmenting ground truth images and by providing segmentation performance as feedback to the synthetic patch generator, they forced the generator to provide samples challenging the segmentation network. Mahmood et al. [12] used a dual Generative Adversarial Network (GAN) that learns to transform masks, including polygons, to synthetic histo-pathological patches.

## EVALUATION OF DEEP LEARNING ARCHITECTURES FOR IMMUNO-FLUORESCENCE BASED NUCLEAR IMAGE SEGMENTATION

We compare the segmentation performance of three well-established architectures and one modification thereof trained to segment nuclear images of immunofluorescently stained samples. The architectures can be divided into two categories: convolutional neural networks (CNNs) (U-Net [10], U-Net with a ResNet34 backbone (U-Net ResNet) and DeepCell [16]) and a CNN with an attached region proposal and segmentation network (Mask R-CNN [23]).

*U-Net:* Similar to autoencoder networks, the U-Net architecture consists of an encoder CNN (contracting path) coupled with a decoder CNN (expansive path) forming a U-like shape. Additional skip-connections between different layers in the down- and up- sampling part of the network are introduced. This strategy allows to preserve spatial information and image details that would otherwise get lost during the down sampling process. We use a theano/lasagne based implementation of the U-Net.

*U-Net ResNet:* The network depth is crucial for image classification and segmentation tasks. An increasing feature representation depth using more feature encoding layers leads to better classification- and segmentation accuracies [24], [25]. When substituting the deconvolution part of the U-Net architecture by a deeper network structure and by using pre-trained weights from e. g. the ImageNet dataset [8], one could expect to increase segmentation accuracy. Therefore, we use a U-Net architecture where the feature encoding part, called the "backbone", was substituted by a ResNet34 [24] architecture, using 34 layers for feature encoding but keeping the skip-connections to ensure spatial resolution for the up-sampling part of the architecture. We use an available keras/tensorflow implementation[1], but changed the loss function to implement the weighted cross-entropy loss setting a higher loss to nuclear borders as suggested by the U-Net authors [10].

*Mask R-CNN:* Mask R-CNN was designed to solve instance aware semantic segmentation problems. The architecture builds upon the Faster Region-based CNN (R-CNN) approach [26] by predicting object masks in parallel to classification of objects in bounding boxes [23]. Thus, unlike the behaviour of DCNNs, Mask R-CNN does not provide a pixel to pixel mapping but rather splits the problem of image segmentation into region detection and subsequent classification and segmentation of region proposals. By focusing on region proposals using coarse spatial quantitation for feature extraction, candidate regions can be extracted with high accuracy. Moreover, constraining the image segmentation to regions within locally extracted bounding boxes may lead to an increased segmentation accuracy when compared to conventional DCNN outputs. We use an available keras/tensorflow implementation of Mask R-CNN[2].

---

[1] https://github.com/qubvel/segmentation_models

*DeepCell:* VanValen et al. [16] proposed a DCNN (Deep-Cell) for cell nuclei segmentation in phase microscopy images of E. coli., fluorescent images of mammalian cell nuclei and phase contrast images. In general a windowing approach is used to generate training candidate images out of the full images, where the window size roughly corresponds to the cell size. A Theano implementation[3] of DeepCell is publicly available.

### Dataset description

We use a recently published dataset [6] consisting of 53 images of IF stained nuclei images containing 3426 nuclei in total. The images are from human ganglioneuroma (GN) tumors (5 images/1807 nuclei), human neuroblastoma (NB) (16 images/700 nuclei) and a human keratinocyte cell line (HaCaT) (32 images/919 nuclei). Among those are GN tissue cryosections (5 images/1807 nuclei), HaCaT cell line cytospin preparations (20 images/645 nuclei), HaCaT cells grown on slide (12 images/274 nuclei), NB bone marrow cytospin preparations (8 images/390 nuclei) and NB touch imprints (8 images/310 nuclei). Automated image acquisition was done using an Axioplan II epifluorescence microscope (Zeiss) equipped with a motorized stage (Maerzhaeuser, Germany) using the Metafer Software (V 3.8.6., Metasystems, Germany). HaCaT and NB samples were acquired with a 63x magnifying objective whereas ganglioneuroma samples were acquired using a 10x magnifying objective. Trained undergraduate students created nuclear label masks using a recently developed machine learning framework [27]. They annotated all images except two. This dataset forms the silver-standard dataset. The two images, derived from NB touch imprints and presenting highly aggregated nuclei, were annotated by an expert using Adobe Illustrator. They were added to the dataset at a later stage to complete it for evaluation of architectures on this type of sample preparation. All annotated images were then carefully curated by a biology expert under the supervision of a pathologist and sent to an external pathologist for review. Finally, all reviewed cases were discussed and nuclear label masks were curated accordingly. The final dataset is denoted as *gold-standard* annotations. Furthermore, we rated each of the images with a score representing a subjective challenge for separating each instance of a nucleus ranging from 1 (almost no challenge) up to 3 (highly challenging image presenting densely packed and aggregated nuclei and/or weakly presented nuclei and/or nuclei with unusual morphology), allowing in-between scores (1-2 and 2-3). The score was set by three experts and, a mean was calculated and rounded to represent one of the three classes.

When training undergraduate students in creating *silver-standard* annotations, experts guided them to annotate only intact nuclei or nuclei where more than 50% of the nucleus

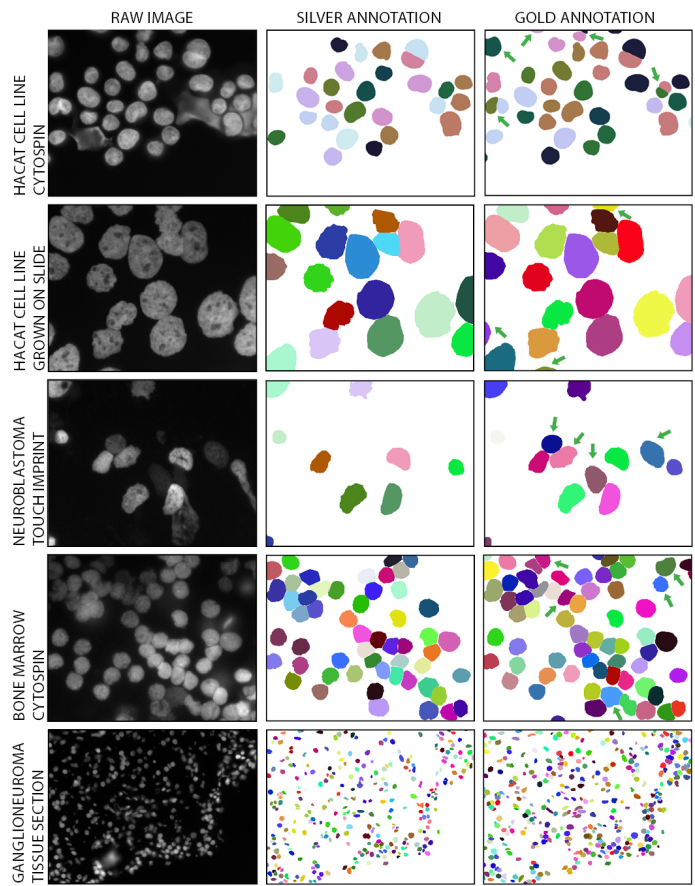

Fig. 2. Examples of all types of preparations/specimen including a comparison between *silver-standard* and *gold-standard* annotations. Green arrows indicate differences between *silver-standard* and *gold-standard* annotations

is visible, as occurs at image borders. Moreover, they were guided to not annotate nuclei with low intensities as such objects would anyhow be excluded from a quantitative analysis of IF signals. Nevertheless, experiments showed that segmentation results of a U-Net trained with the silver-standard dataset were not satisfying. Thus, we decided to include and annotate all visible and intact nuclei in *gold-standard* annotations. Examples of all types of preparations/specimens including a comparison between *silver-standard* and *gold-standard* annotations are given in Figure 2[4]. The green arrows depicted in Figure 2 indicate differences between *silver-* and *gold-standard* annotations. It can be observed that nuclei with weak intensity and nuclei partially present at image borders are annotated in the *gold-standard* annotations when compared to the *silver-standard* annotations, leading to an increased number of annotated nuclei especially in GN tissue cryosections.

### Artificial image generation

We evaluate the influence of adding artificially generated images to the training and validation set on segmentation performance. In contrast to Hou et al. [22], we do not create artificial nuclei and place them on nuclei-free background patches, we rather create artificial images by modelling a realistic background and by placing cropped and augmented nuclei on this artificially created background. Therefore, we can create artificial images using individually augmented nuclei, addressing the varying nuclei intensities. Moreover, as discussed in the introduction, the overall aim of IF based nuclear image segmentation within this work is to separate each nucleus to ensure instance aware segmentation. Thus, we use a strategy to create artificial images with highly overlapping nuclei, forcing the deep learning architectures trained on these images to learn how to split aggregated nuclei. The strategy to create artificial images is as follows: we randomly select an annotated natural image and the corresponding annotation mask from one of the datasets given a certain tissue origin/diagnosis. By dilating the foreground regions of the binarized annotation mask using a circle-shaped structuring element of size 15, pixels with values higher than the mean background intensity, occurring due to blurred nuclei, are included. When inverting the resulting mask, only the region of pixels representing the background signal is covered. We now iteratively sample the intensity values of random pixels from this region and assign them to one of the pixels of an empty image patch. This is done until all pixels of the respective image patch are set. Thus, the background pixel value distribution of the created image patch roughly matches the one of the original image. Subsequently, we use arbitrary nuclei cropped from the annotated image, transform them by rotation, translation, elastic deformation, intensity variation and combinations of those operations. The transformed, cropped nuclei are then placed at crossing positions of a grid virtually overlaid with the image patch, including a randomly added offset in x- and y- direction. The maximum offset value depicts the probability of a nucleus to overlap with a neighbor nucleus placed on the grid. The larger the offset, the higher the probability to overlap with a neighbor nucleus. For each crossing position on the grid, we randomly decide if the object shall be placed or not. If a nucleus to be inserted overlaps another nucleus already present, we randomly decide if the overlapping part of the nucleus replaces the overlapping part of the formerly present nucleus, or if the part is added to the part of the formerly present nucleus multiplied with a constant value between 0 and 1 to imitate overlap. Thus, we can simulate aggregating and overlapping nuclei. The same augmentation and placement is done for cropped nuclear masks, placed on a new image mask patch, except that placed nuclei masks always replace overlapping parts of existing nuclei masks as we do not model masks with fuzzy annotations.

---

[2]https://github.com/matterport/Mask_RCNN
[3]https://github.com/CovertLab/DeepCell

[4]A detailed comparison can be obtained in Suppl. Table 1

Finally, we obtain a nuclear and a mask image. The first contains random nuclei augmented with arbitrary image transformation strategies. The latter contains labeled objects that were placed on the very same positions as the nuclei in the nuclear image.

Nevertheless, we observed that training DL architectures using such artificially generated images does not lead to better segmentation results. This may be due to the fact that nuclei naturally showing blurred borders in IF images do not show those when cropped, transformed and placed on new image patches, guiding the network to learn features differently from natural image features. To overcome this issue, we trained an image-to-image translation GAN [28] called pix2pix to learn the transformation of artificially generated images into natural-like images. The GAN is trained on pairs of natural and artificial images, where nuclei are cropped from natural image patches and placed at the very same position on a new image patch[5]. By training the network on these paired images, the network is forced to learn the implicit transformation of artificial to natural-like images. The final workflow to create natural-like artificial images is depicted in Figure 3.

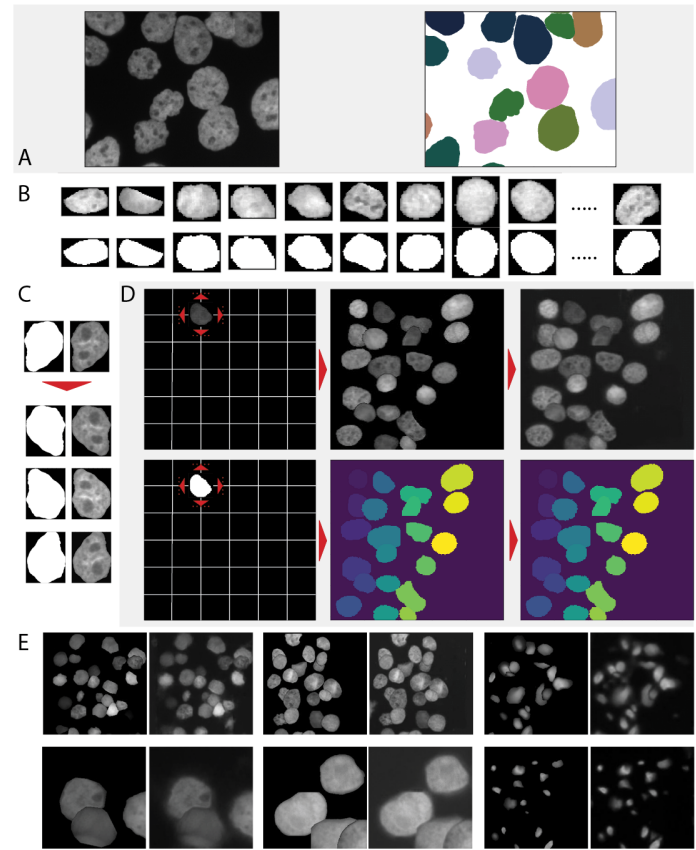

Fig. 3. Generation of natural-like artificial image patches. A. Nuclei and respective mask objects of a training image patch are collected and B. cropped from the raw nuclei image respective mask. C. Each nucleus patch is arbitrarily augmented using rotation, intensity variation, elastic deformation, flipping, etc. The same morphological transformations are applied to the mask patch. D. We create a new image patch and set the background pixels sampled from the original raw image background. Subsequently, we place nuclei at certain positions induced by a grid sized 3x3, 5x5, etc. For each position on the grid, we randomly decide if a nucleus shall be placed there and if so, we add a random offset, where the maximum offset indicates the probability to overlap with neighbor nuclei. The same placement is performed for mask objects on a new mask patch. Finally, a GAN is used to transform the artificial image into a natural-like image. This is done for each dataset independently. E. Randomly created patches. Upper row: images were scaled such that all nuclei have the same mean size. Lower row: no scaling was applied. First two columns: artificial/natural-like HaCaT image. Third and fourth column: artificial/natural-like neuroblastoma image. Last two columns: artificial/natural-like ganglioneuroma image.

### Pipeline for architecture comparison

To evaluate the potential of state-of-the-art architectures to segment nuclear images across various tissue origins and sample preparation types without the need to apply sample specific post-processing steps, we set up a pipeline to enable an objective comparison. The code is publicly available[6]. The pipeline as illustrated in Figure 4 operates as follows. We split the annotated *gold-standard* dataset into training, validation and test set, for all of the three different tissue origins (HaCaT cellline, NB tumor, GN tumor) separately. We do not consider the different preparation types applied to the imaged samples for architecture training, but we split the dataset such that at least one image of each sample preparation type is contained in the training, validation and test set. All images of these datasets are further called natural images, in contrast to natural-like artificial images that result from artificial image synthesis. We set the input layer size of all deep learning architectures to the size 256x256[7]. To prepare the dataset to fit the given input layer size, we apply a tiling strategy to the dataset images as proposed by Ronneberger et al. [10] in order to obtain image patches sized 256x256. By using this strategy we observe the following benefits: 1) The training and validation set images are extended. 2) Overlapping tiles prevent artifacts at tile borders when reconstructing final network predictions after network inference on the test set image tiles. Moreover, we rescale the natural images such that nuclei are equally mean-sized across all images as we want to evaluate the impact of rescaling images on the segmentation performance. To rescale images, the mean nuclear size of an image was calculated based on the mean size of all nuclear mask objects within the corresponding mask image. Subsequently, all images and masks were resized such that nuclei have the same mean size per image across all images. In addition to the natural images of the training and validation set, we create artificial images of size 256x256 as described in Section *Artificial image generation* and add them to the training- or validation set, respectively. As we aim to additionally compare the segmentation performance for all architectures between *silver-standard* and *gold-standard* training sets, we apply the same steps except for generating artificial images to the *silver-standard* dataset.

---

[5]see Suppl. Figure 1

We then train all architectures four times using

- the non-scaled natural images of the *gold-standard* training set
- the scaled natural images of the *gold-standard* training set containing equally mean-sized nuclei across images
- the scaled natural images of the *silver-standard* training set containing equally mean-sized nuclei across images
- the scaled natural and equally scaled natural-like artificial images of the *gold-standard* training set

After network inference on the test set patches, the patches are reassembled and rescaled to fit the original image size. Thus, prediction results can be compared across all architectures. The output of the tested networks differs: while CNNs, utilized for image segmentation, result in a probability map, Mask R-CNN inference result in object masks, one mask for each detected object. We threshold the resulting, reassembled probability maps of the purely convolutional architectures by a value of 0.5 to obtain binary masks and label them to obtain the labeled object masks[8]. For reassembled Mask R-CNN predictions, we label each object and add it to a new image mask to obtain the final labeled object mask. The only post-processing step we apply is to remove small artifacts that do not fit the size of nuclei, where the threshold depends on the image magnification. This post-processing is independent of shape or intensity-based features and is not specifc to a certain sample preparation type or tissue type. Contours of predicted objects can slightly vary between the U-Net and the other architectures. This is because the U-Net architecture is trained with binary images where all pixels are weighted in addition, whereas DeepCell and Mask R-CNN are trained with labeled masks. To force the U-Net architecture to learn how to split nuclear aggregations, highest weights are set for regions between touching nuclei. To transform originally labeled masks into binary masks where nuclear borders are represented as background pixels, nuclear objects within masks have to be shrunk by a morphological erosion operation. Thus, segmented objects predicted by the U-Net architecture are systematically smaller than objects predicted by Mask R-CNN or DeepCell, influencing pixel-level evaluation scores. This does not depict a problem as we focus the evaluation on instance aware segmentation results rather than on observing exact matches between predicted objects and groundtruth.

### Evaluation metrics

Most authors provide object-level as well as pixel-level metrics to evaluate nuclear segmentation methods. To overcome the problem of choosing the right class of metrics, Kumar et al. [29] proposed a combined metric, the Aggregated Jaccard Index (AJI), taking object- and pixel-level errors into account. The AJI is prominently used in recent publications, see [14], [30], [12] for examples. The disadvantage of using this combined metric is that it is not obvious if pixel or object-level errors contribute to a low AJI score.

We focus our research aims on quantification of nuclear bound antibody/Flourescence in situ hybridisation signals for subsequent classification of cellular types and comparison of multiple experimental conditions. Therefore, we evaluate all architectures using the object-level metrics precision and recall and provide US rate and the AJI score in addition.

We consider a ground truth object to be detected if more than 50% of the groundtruth object's pixels are covered by predictions, and assign it as FN otherwise. We count a ground truth object as TP, if it is overlapped by exactly one predicted object with a JI between ground truth object and predicted object of greater than 0.5. Predicted objects only touching a part of the object would count as FP. If more than one predicted object overlaps the ground truth object such that the overlapping area covers more than 50% of the predicted object's area, the ground truth object is considered as OS. An object is classified as FP, if it overlaps with the ground truth for less than 50%[9]. If more than one ground truth object overlaps the predicted object such that the overlapping area covers more than 50% of the ground truth object's area, the ground truth objects overlapped by the prediction are classified as US. We report US as the ratio between the number of under-segmented nuclei and the number of ground truth objects.

---

[8]see Suppl. Figure 2

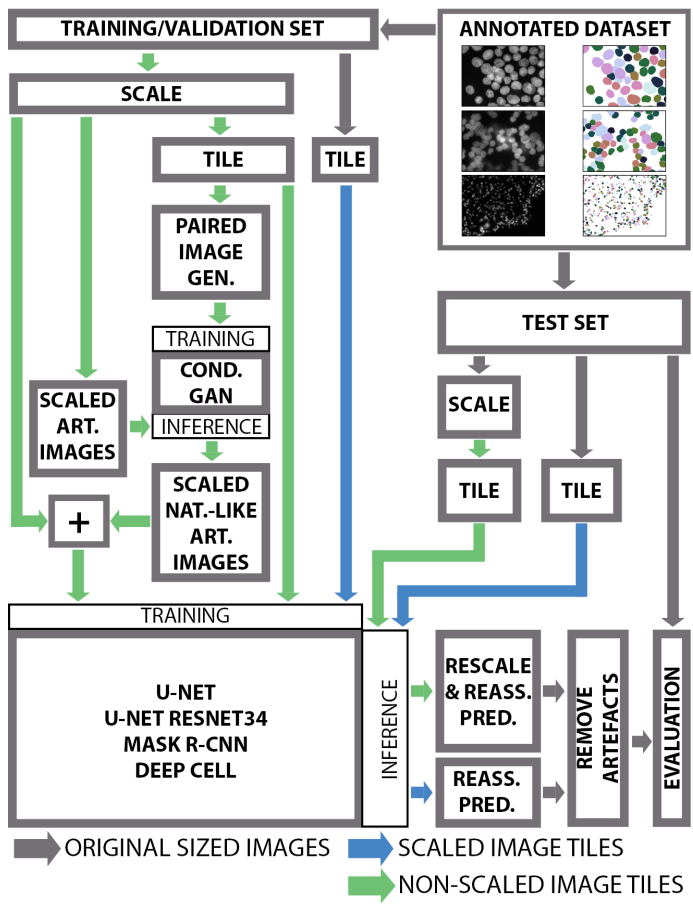

**Fig. 4.** Pipeline for training and evaluation of deep learning architectures for instance aware nuclei image segmentation.

## Results

Segmentation performance of deep learning architectures depends on their design and on several additional factors such as the size of the dataset and the architectures' hyper-parameters. To evaluate the influence of additional conditions such as different tissue origins, preparation types, annotation quality, nuclear scales, artificial images and varying segmentation complexity levels, we fixed the hyperparameters of all architectures[10] and trained them with the images generated at these conditions. We report object level metrics, namely precision, recall and US [11]. We do not report OS since there is almost no OS occurring. In this work, we focus on instance aware segmentation results rather than on observing exact matches between predicted objects and ground truth. As discussed in the previous section, pixel-level metrics are

[9]Examples of possible cases of ground truth objects and predictions are illustrated in Suppl. Figure 3

[10]see Suppl. Table 2

[11]Qualitative results are presented in Suppl. Figure 4

not reliable to compare between architectures in this study. Nonetheless, we report the AJI score to show a comparison between the aforementioned differing sample conditions on the same architecture.

### B. The influence of scaling

We observed that image segmentation using small multi-scale image datasets (multi-levels of magnification) can lead to inaccurate segmentation results. This is because in fluorescence nuclear images, nuclei acquired using a 10x magnifying objective can have similar shape and texture as spot-like inclusions in nuclei acquired using a 63x magnifying objective. We hypothesize that rescaling all images such that the mean nuclear size is equal across all images will lead to better segmentation results. The influence of scaling vs. non-scaling is presented in Table I.

| Metric | U-Net<br>NON/SCL | U-Net ResNet<br>NON/SCL | MRCNN<br>NON/SCL | DeepCell<br>NON/SCL |
|---|---|---|---|---|
| US | 0.29/0.04 | 0.19/0.16 | 0.09/**0.02** | 0.41/0.42 |
| REC | 0.64/0.85 | 0.77/0.77 | 0.83/**0.87** | 0.50/0.48 |
| PREC | 0.79/0.87 | 0.86/0.84 | 0.89/**0.91** | 0.74/0.74 |
| AJI | 0.66/**0.75** | 0.74/0.72 | 0.73/0.69 | 0.55/0.53 |

TABLE I
EVALUATION METRICS CONSIDERING SCALED VS. NON-SCALED IMAGES ON THE GOLD-STANDARD TEST SET. BEST METRIC IS HIGHLIGHTED.

Rescaling all images to the same mean nuclei size leads to an increased recall and precision using the U-Net and the Mask R-CNN architecture, whereas the U-Net with ResNet34 backbone and the DeepCell architecture metrics are slightly worse. US decreases for all architectures except for DeepCell. The AJI score slightly decreases for all architectures except for the U-Net when using scaled images.

### C. Silver vs. gold-standard training

As discussed in Section *Dataset description*, two types of annotated datasets were generated: *silver* and *gold-standard* annotations. We trained all architectures on both datasets using scaled images. The results are presented in Table II.

| Metric | U-Net<br>silv/gold | U-Net ResNet<br>silv/gold | MRCNN<br>silv/gold | DeepCell<br>silv/gold |
|---|---|---|---|---|
| US | **0.02**/0.04 | 0.03/0.16 | **0.02**/**0.02** | 0.36/0.42 |
| REC | 0.80/0.85 | 0.75/0.77 | **0.90**/0.87 | 0.58/0.48 |
| PREC | 0.78/0.87 | 0.79/0.84 | 0.81/**0.91** | 0.68/0.74 |
| AJI | 0.71/**0.75** | 0.71/0.72 | 0.65/0.69 | 0.56/0.53 |

TABLE II
SILVER- VS GOLD-STANDARD TRAINING, EVALUATED ON SCALED IMAGES OF THE GOLD-STANDARD TEST SET. BEST METRIC IS HIGHLIGHTED.

The results show an improved precision for all architectures except for DeepCell when trained on *gold-standard* images. Recall only increases for the U-Net and U-Net ResNet34 architectures, but decreases for Mask R-CNN and DeepCell. Taking into account pixel and object-level comparison by considering the AJI score, except for the DeepCell architecture an overall improvement is achieved when training on *gold-standard* images.

## D. Performance on cells of different specimens

We analyze the segmentation performance of all architectures trained on natural scaled nuclei images of three different specimens: HaCaT cell line, neuroblastoma and ganglioneuroma patient samples. To test for the influence of artificially created image data, we additionally trained all architectures on a combination of scaled natural and artificial images. Evaluation metrics using scaled natural images and the combination of scaled natural and artificial images are presented in Table III. Precision and recall for all training sets are visualized in Figure 5 including the results on non-scaled images for comparison.

| Metric | U-Net nat/nat+art | U-Net ResNet nat/nat+art | MRCNN nat/nat+art | DeepCell nat/nat+art |
|---|---|---|---|---|
| Specimen: HaCaT, #images:6, #nuclei:167 | | | | |
| US | 0.04/0.02 | 0.06/0.02 | **0.01/0.01** | 0.39/0.36 |
| REC | 0.92/0.92 | 0.91/**0.94** | 0.94/**0.95** | 0.57/0.59 |
| PREC | 0.94/0.94 | 0.94/0.95 | **0.97/0.97** | 0.82/0.79 |
| AJI | **0.91**/0.88 | 0.89/0.88 | 0.80/0.76 | 0.58/0.62 |
| Specimen: neuroblastoma #images:4, #nuclei:218 | | | | |
| US | 0.03/0.15 | 0.20/0.17 | **0.01**/0.02 | 0.59/0.62 |
| REC | 0.81/0.73 | 0.70/0.80 | 0.83/**0.89** | 0.30/0.31 |
| PREC | 0.83/0.79 | 0.80/0.85 | **0.88**/0.87 | 0.66/0.71 |
| AJI | **0.67**/0.65 | 0.61/0.66 | 0.64/0.66 | 0.38/0.36 |
| Specimen: ganglioneuroma #images:1 #nuclei:365 | | | | |
| US | 0.07/0.15 | 0.21/0.20 | **0.05**/0.12 | 0.28/0.3 |
| REC | 0.81/0.71 | 0.69/0.71 | **0.85**/0.79 | 0.56/0.54 |
| PREC | 0.85/0.77 | 0.79/0.82 | **0.87**/0.78 | 0.73/0.72 |
| AJI | **0.68**/0.60 | 0.67/0.67 | 0.64/0.60 | 0.61/0.57 |

TABLE III
Evaluation on different specimens using scaled images of the gold-standard test set. Best metric is highlighted.

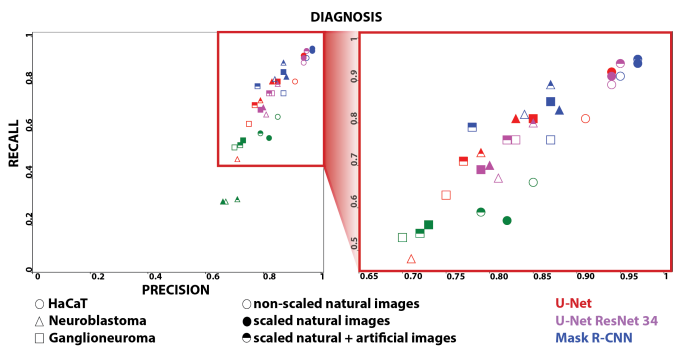

Fig. 5. Performance (precision vs. recall) of architectures trained on samples of different specimens: HaCaT, neuroblastoma and ganglioneuroma.

The results show that the segmentation performance differs between specimens. The benefit of using artificial image data depends on the architecture used. Overall, segmentation of HaCaT cells leads to best recall and precision scores. Mask R-CNN almost always achieves a higher recall than the other architectures, whereas DeepCell achieves lowest overall scores. Furthermore, adding artificial training data leads to an increased recall in HaCaT and neuroblastoma images, whereas recall decreases for the U-Net. Precision is not affected by adding artificial images for HaCaT cells, decreases for ganglioneuroma samples and varies among architectures in neuroblastoma samples. Scaled images lead to an overall higher recall and precision when compared to non-scaled images across all specimens, except when using the DeepCell architecture. The benefit of adding artificial image data is highest for the U-Net Resnet34 architecture, where recall and precision improve in all specimens.

## E. Performance on different types of preparation

We analyze the segmentation performance of all architectures on five different types of sample preparation: bone marrow cytospins, cell line cytospins, cells grown on slide, tumor touch imprints and tissue cryosections. To show the benefit of artificially created image data, we trained all architectures on natural and additionally on a combination of natural and artificial images. The results are presented in Table IV. Precision and recall are visualized in Figure 6 including the results for non-scaled images for comparison.

| Metric | U-Net nat/nat+art | U-Net ResNet nat/nat+art | MRCNN nat/nat+art | DeepCell nat/nat+art |
|---|---|---|---|---|
| Preparation: cell line cytospin, #images:5, #nuclei:149 | | | | |
| US | 0.04/0.03 | 0.07/0.03 | **0.01/0.01** | 0.39/0.38 |
| REC | 0.91/0.92 | 0.91/0.94 | 0.94/**0.95** | 0.58/0.59 |
| PREC | 0.94/0.94 | 0.94/0.95 | **0.97/0.97** | 0.83/0.79 |
| AJI | **0.90**/0.88 | 0.88/0.86 | 0.79/0.75 | 0.57/0.61 |
| Preparation: cell line grown, #images:1, #nuclei:18 | | | | |
| US | **0/0** | **0/0** | **0/0** | 0.39/0.22 |
| REC | **0.94**/0.89 | **0.94/0.94** | 0.94/0.94 | 0.50/0.61 |
| PREC | **1**/0.94 | **1/1** | **1/1** | 0.75/0.79 |
| AJI | **0.96**/0.92 | **0.96**/0.95 | 0.83/0.82 | 0.61/0.70 |
| Preparation: bone marrow cytospin, #images:1, #nuclei:124 | | | | |
| US | **0.02**/0.23 | 0.32/0.26 | **0.02/0.02** | 0.84/0.90 |
| REC | 0.75/0.63 | 0.54/0.71 | 0.77/**0.85** | 0.09/0.08 |
| PREC | 0.77/0.72 | 0.68/0.81 | 0.83/**0.85** | 0.37/0.38 |
| AJI | 0.55/0.56 | 0.42/0.56 | 0.55/**0.62** | 0.19/0.16 |
| Preparation: tumor touch imprint, #images:3, #nuclei:94 | | | | |
| US | 0.04/0.04 | 0.04/0.04 | **0**/0.02 | 0.27/0.26 |
| REC | 0.88/0.86 | 0.90/0.93 | 0.93/**0.94** | 0.57/0.61 |
| PREC | 0.90/0.87 | 0.92/0.89 | **0.96**/0.90 | 0.79/0.83 |
| AJI | 0.81/0.76 | **0.83**/0.73 | 0.74/0.70 | 0.58/0.57 |

TABLE IV
Evaluation on different types of preparations, trained using scaled images of the gold-standard training set. Results for tissue cryosections are similar to ganglioneuroma results (see Table III).

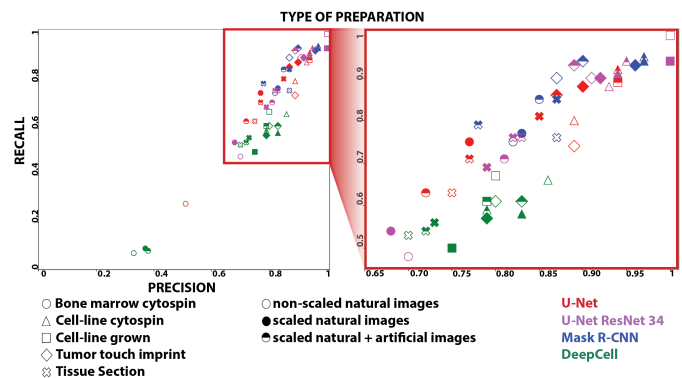

Fig. 6. Performance (precision vs. recall) of architectures for different types of sample preparation.

Mask R-CNN achieves the best overall precision and recall values, whereas Deep cell achieves lowest overall scores.

While grown and cytospinned cell lines can be segmented with high precision and recall, tissue cryosections and bone marrow cytospins are most challenging to segment. This is based on the observation that those preparation methods contain a lot of aggregated cells, challenging the architectures to separate those. Using Mask R-CNN, the use of artificial image data leads to an increased recall but decrease in precision, whereas using the U-Net with ResNet34 backbone trained on scaled natural and artificial image data leads to an increase in recall and precision in bone marrow cytospins and cell line cytospins.

### F. Influence on different segmentation challenge levels

We observed a variation of segmentation performance between different architectures, different specimens and image scaling. To give an objective rating of architecture recommendation, we aimed at removing the varying parameters. To do so, we created a classification reflecting the level of challenge to segment an image as described in Section *Dataset description*. The results are shown in Table V and visualized in Figure 7.

| Metric | U-Net nat/nat+art | U-Net ResNet nat/nat+art | MRCNN nat/nat+art | DeepCell nat/nat+art |
|---|---|---|---|---|
| Challenge: low level, #images:6, #nuclei:220 | | | | |
| US | 0.04/0.03 | 0.05/0.03 | **0.01**/0.02 | 0.27/0.25 |
| REC | 0.91/0.90 | 0.91/**0.94** | 0.93/**0.94** | 0.63/0.66 |
| PREC | 0.93/0.92 | 0.94/0.93 | **0.96**/0.94 | 0.83/0.83 |
| AJI | **0.89**/0.86 | **0.89**/0.85 | 0.79/0.75 | 0.65/0.69 |
| Challenge: medium level, #images:2, #nuclei:165 | | | | |
| US | 0.02/0.19 | 0.27/0.21 | **0.01/0.01** | 0.81/0.86 |
| REC | 0.78/0.69 | 0.62/0.76 | 0.81/**0.88** | 0.13/0.13 |
| PREC | 0.80/0.77 | 0.75/0.84 | 0.86/**0.88** | 0.77/0.45 |
| AJI | **0.68/0.68** | 0.60/**0.68** | 0.64/0.67 | 0.27/0.25 |

TABLE V

EVALUATION ON DIFFERENT TYPES OF CHALLENGE LEVELS, TRAINED USING THE GOLD-STANDARD TRAINING SET AND SCALED IMAGES. RESULTS OF HIGH SEGMENTATION CHALLENGE CORRESPOND TO GANGLIONEUROMA RESULTS AND CAN BE OBSERVED IN TABLE III.

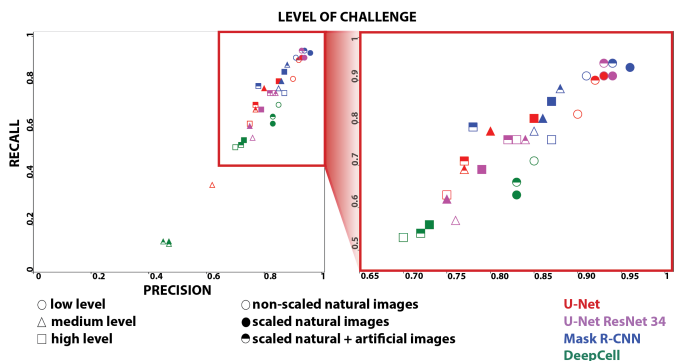

Fig. 7. Performance (precision vs. recall) of architectures for different types of segmentation challenge levels, trained using scaled images of the *gold-standard* test set.

Images presenting low challenge for instance aware segmentation achieve the highest scores as expected, whereas highly challenging images achieve the lowest scores except for DeepCell achieving the lowest scores for medium challenging images. Mask R-CNN achieves the best overall recall scores. The use of artificial images is of highest benefit for the U-Net

ResNet. For medium challenging images, Mask R-CNN recall and precision scores are increased with artificial training data, whereas precision is decreased for highly challenging images.

### DISCUSSION

Quantitative image analysis using deep learning based segmentation approaches enables researchers to detect even subtle biological effects if the utilized segmentation architectures allow for an instance aware nuclear segmentation with high accuracy. We compared and evaluated the segmentation performance of four deep learning architectures, trained and tested on fluorescent nuclear images of different specimen, sample preparation types, annotation quality, image scales and segmentation complexity levels.

Results demonstrate that rescaling images to obtain equally mean-sized nuclei across images increases precision and recall scores. We hypothesize that using images at the same scale level reduces the segmentation complexity, as architectures are no longer forced to learn image features along multiple scale levels. The influence of ground truth annotation quality is expressed by the precision score: carefully annotated, consistent annotations increase the precision, whereas the influence on recall depends on the architecture used. When comparing different specimens and types of preparation, Mask R-CNN overall performs the best by a large margin, whereas DeepCell performs the worst. Mask R-CNN's architecture design to separate object detection and object segmentation results in high precision and recall scores, independent of the specimen analyzed or the type of preparation. The advantage of using artificial training data can be observed especially in the case of HaCaT and neuroblastoma specimen segmentation. The impact is highest for segmenting bone marrow cytospin images, whereas a negative effect appears when segmenting ganglioneuroma tissue cryosections. This might be either due to the 10x image magnification used for capturing the tissue cryosections, the preparation of tissue cryosections or the complex morphology of the cells present in sections in general. Low magnification, however, leads to decreased texture details, which is known to be an issue in ImageNet pre-trained CNNs [31]. When analyzing the results for the different levels of segmentation challenge, it can be observed that the use of artificial images is most beneficial for medium challenging images. One explanation of this effect is that these images are captured with a 63x magnifying objective presenting detailed texture, high variation of nuclei intensities and also highly aggregated or overlapping nuclei. The added benefit when using artificial images is expected as these conditions are best simulated by the artificial training data generated. In low challenging images, nuclei are not aggregated, thus artificial images cannot contribute to increasing the segmentation performance.

### CONCLUSION

Fluorescent nuclear images across different tissue origins, sample preparation types and image magnification, can be segmented by three out of the four evaluated deep learning architectures (U-Net, U-Net ResNet, Mask R-CNN) with a satisfactory segmentation performance. Moreover, we provide evidence that the segmentation performance is dependent on the following factors: 1. *The type of framework used.* Mask R-CNN outperforms all other frameworks when evaluated across images of all types of tissue origin and sample preparation. 2. *The type of sample preparation used.* Cell line cytospins and cell lines grown on microscopy glass slides can be segmented with highest precision and recall, tissue cryosections with the lowest. 3. *The quality of the annotated dataset.* The more accurate the nuclear annotations are, the higher precision and recall are. An exception is depicted by Mask R-CNN, which is less sensitive to inaccurate nuclear annotations. 3. The use of artificially created images and annotations. We show that by extending the dataset with artificially created images and masks, evaluation scores increase, except for nuclear images of tissue cryosections. 4. *Rescaling of nuclear images.* We discovered that by rescaling nuclear images and masks to the same mean nuclear size across all images, evaluation metric values increase. 5. *The complexity of a nuclear image depicted by the grade of nuclei overlap/nuclear aggregation and the variance of staining intensity.* Images presenting low segmentation complexity can by segmented with high precision and recall. For images presenting medium- and high segmentation complexity, precision and recall depend on the architecture used and on the use of artificial images.

Considering the recent revival and importance of single cell analysis in biomedical research, this work has the potential to improve the accuracy and enable broad application of fluorescence microscopy based image analysis workflows.

## ACKNOWLEDGMENT

This work was carried out within the Austrian Research Promotion Agency (FFG) COIN Networks projects TISQUANT and VISIOMICS and additionally supported by the Austrian Ministry for Transport, Innovation and Technology, the Federal Ministry of Science, Research and Economy, the Province of Upper Austria in the frame of the COMET center SCCH and the St. Anna Kinderkrebsforschung.

## REFERENCES

[1] K. Huang and R. F. Murphy, "From Quantitative Microscopy to Automated Image Understanding," *Journal of Biomedical Optics*, vol. 9, no. 5, pp. 893–912, 2004.

[2] I. M. Ambros, J. I. Hata, V. V. Joshi, B. Roald, L. P. Dehner, H. Tüchler, U. Pötschger, and H. Shimada, "Morphologic features of neuroblastoma (Schwannian stroma-poor tumors) in clinically favorable and unfavorable groups," *Cancer*, vol. 94, no. 5, pp. 1574–1583, 2002.

[3] M. N. Gurcan, L. E. Boucheron, A. Can, A. Madabhushi, N. M. Rajpoot, and B. Yener, "Histopathological Image Analysis: A Review," *Biomedical Engineering, IEEE Reviews in*, vol. 2, pp. 147–171, 2009.

[4] O. Ronneberger, D. Baddeley, F. Scheipl, P. J. Verveer, H. Burkhardt, C. Cremer, L. Fahrmeir, T. Cremer, and B. Joffe, "Spatial quantitative analysis of fluorescently labeled nuclear structures: Problems, methods, pitfalls," *Chromosome Research*, vol. 16, pp. 523–562, 2008.

[5] A. A. Hill, P. LaPan, Y. Li, and S. Haney, "Impact of image segmentation on high-content screening data quality for SK-BR-3 cells," *BMC Bioinformatics*, vol. 8, pp. 1–13, 2007.

[6] F. Kromp, "An annotated fluorescence image dataset for training nuclear segmentation methods," 2019. [Online]. Available: https://identifiers.org/biostudies:S-BSST265

[7] F. Xing and L. Yang, "Robust Nucleus/Cell Detection and Segmentation in Digital Pathology and Microscopy Images: A Comprehensive Review," *iEEE Rev Biomed Eng.*, vol. 9, pp. 234–263, 2016.

[8] A. Krizhevsky, I. Sutskever, and G. E. Hinton, "ImageNet Classification with Deep Convolutional Neural Networks," *Advances In Neural Information Processing Systems*, pp. 1–9, 2012.

[9] W. Zhang, R. Li, H. Deng, L. Wang, W. Lin, S. Ji, and D. Shen, "Deep Convolutional Neural Networks for Multi-Modality Isointense Infant Brain Image Segmentation," *NeuroImage*, vol. 108, pp. 214–224, 2016.

[10] O. Ronneberger, P. Fischer, and T. Brox, "U-net: Convolutional networks for biomedical image segmentation," in *Medical Image Computing and Computer-Assisted Intervention – MICCAI 2015*, 2015, pp. 234–241.

[11] Y. Cui, G. Zhang, Z. Liu, Z. Xiong, and J. Hu, "A Deep Learning Algorithm for One-step Contour Aware Nuclei Segmentation of Histopathological Images," *arXiv preprint*, pp. 1–13, 2018.

[12] F. Mahmood, D. Borders, R. Chen, G. N. Mckay, K. J. Salimian, A. Baras, and N. J. Durr, "Deep Adversarial Training for Multi-Organ Nuclei Segmentation in Histopathology Images," *arXiv preprint*, 2018.

[13] Z. Alom, C. Yakopcic, T. M. Taha, and V. K. Asari, "Nuclei Segmentation with Recurrent Residual Convolutional Neural Networks based U-Net," *IEEE NAECON 2018*, pp. 228–233, 2018.

[14] P. Naylor, M. Lae, F. Reyal, and T. Walter, "Segmentation of Nuclei in Histopathology Images by deep regression of the distance map," *IEEE Transactions on Medical Imaging*, vol. 38, no. 2, pp. 448–459, 2019.

[15] J. Caicedo, J. Roth, A. Goodman, T. Becker, K. W. Karrhois, C. McQuin, S. Singh, and A. E. Carpenter, "Evaluation of Deep Learning Strategies for Nucleus Segmentation in Fluorescence Images," *bioRxiv*, 2018.

[16] D. A. Van Valen, T. Kudo, K. M. Lane, D. N. Macklin, N. T. Quach, M. M. DeFelice, I. Maayan, Y. Tanouchi, E. A. Ashley, and M. W. Covert, "Deep Learning Automates the Quantitative Analysis of Individual Cells in Live-Cell Imaging Experiments," *PLoS Computational Biology*, vol. 12, no. 11, pp. 1–24, 2016.

[17] D. J. Ho, C. Fu, P. Salama, K. W. Dunn, and E. J. Delp, "Nuclei Segmentation of Fluorescence Microscopy Images Using Three Dimensional Convolutional Neural Networks," *2017 IEEE Conference on Computer Vision and Pattern Recognition Workshops*, no. 1, pp. 834–842, 2017.

[18] A. Kendall, V. Badrinarayanan, and R. Cipolla, "Bayesian SegNet: Model Uncertainty in Deep Convolutional Encoder-Decoder Architectures for Scene Understanding," *arXiv preprint*, 2015.

[19] S. K. Sadanandan, P. Ranefall, S. Le Guyader, and C. Wählby, "Automated Training of Deep Convolutional Neural Networks for Cell Segmentation," *Nature Scientific Reports*, vol. 7, no. 1, pp. 1–7, 2017.

[20] J. Shijie, W. Ping, J. Peiyi, and H. Siping, "Research on data augmentation for image classification based on convolution neural networks," *Proceedings - 2017 Chinese Automation Congress, CAC 2017*, vol. 2017-Janua, no. 201602118, pp. 4165–4170, 2017.

[21] R. A. Russell, N. M. Adams, D. A. Stephens, E. Batty, K. Jensen, and P. S. Freemont, "Segmentation of fluorescence microscopy images for quantitative analysis of cell nuclear architecture," *Biophysical Journal*, vol. 96, no. 8, pp. 3379–3389, 2009.

[22] L. Hou, A. Agarwal, D. Samaras, T. M. Kurc, R. R. Gupta, and J. H. Saltz, "Unsupervised Histopathology Image Synthesis," *arXiv*, 2017.

[23] S. Ren, K. He, R. Girshick, and J. Sun, "Faster R-CNN: Towards Real-Time Object Detection with Region Proposal Networks," *IEEE TPAMI*, vol. 39, no. 6, pp. 1137–1149, 2017.

[24] S. Wu, S. Zhong, and Y. Liu, "Deep residual learning for image steganalysis," *Multimedia Tools and Applications*, pp. 1–17, 2017.

[25] K. Simonyan and A. Zisserman, "Very Deep Convolutional Networks for large-scale image recognition," *ICLR 2015*, 2015.

[26] K. He, G. Gkioxari, P. Dollár, and R. Girshick, "Mask R-CNN," *IEEE International Conference on Computer Vision*, pp. 2980–2988, 2017.

[27] F. Kromp, I. Ambros, T. Weiss, D. Bogen, H. Dodig, M. Berneder, T. Gerber, S. Taschner-Mandl, P. Ambros, and A. Hanbury, "Machine learning framework incorporating expert knowledge in tissue image annotation," *IEEE ICPR*, pp. 343–348, 2016.

[28] P. Isola, J.-Y. Zhu, T. Zhou, and A. A. Efros, "Image-to-Image Translation with Conditional Adversarial Networks," *arXiv preprint*, 2017.

[29] N. Kumar, R. Verma, S. Sharma, S. Bhargava, A. Vahadane, and A. Sethi, "A Dataset and a Technique for Generalized Nuclear Segmentation for Computational Pathology," *IEEE Transactions on Medical Imaging*, vol. 36, no. 7, pp. 1550–1560, 2017.

[30] Z. Zeng, W. Xie, Y. Zhang, and Y. Lu, "RIC-Unet: An Improved Neural Network Based on Unet for Nuclei Segmentation in Histology Images," *IEEE Access*, vol. 7, pp. 21420–21428, 2019.

[31] R. Geirhos, C. Michaelis, F. Wichmann, P. Rubisch, M. Bethge, and W. Brendel, "ImageNet - trained CNNs are biased towards texture ; increasing shape bias improves accuracy and robustness," *arXiv*, 2018.

# Deep Learning architectures for generalized immunofluorescence based nuclear image segmentation: Supplementary material

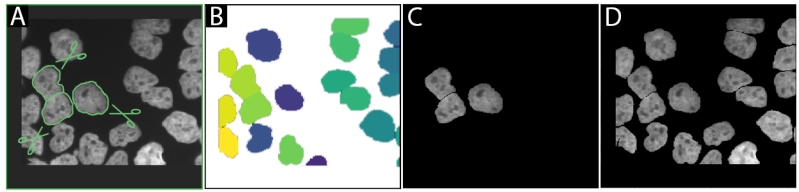

Fig. 1. Creating paired image patches for GAN training. A. Image patch from an original raw nuclear image and B. nuclear mask after applying image augmentation using image flipping and elastic deformations. C. Nuclei from augmented raw nuclei images are cropped and placed into a new artificial image. D. Created artificial nuclear image (D) and corresponding augmented natural nuclear image (A) form an image pair serving for training the GAN.

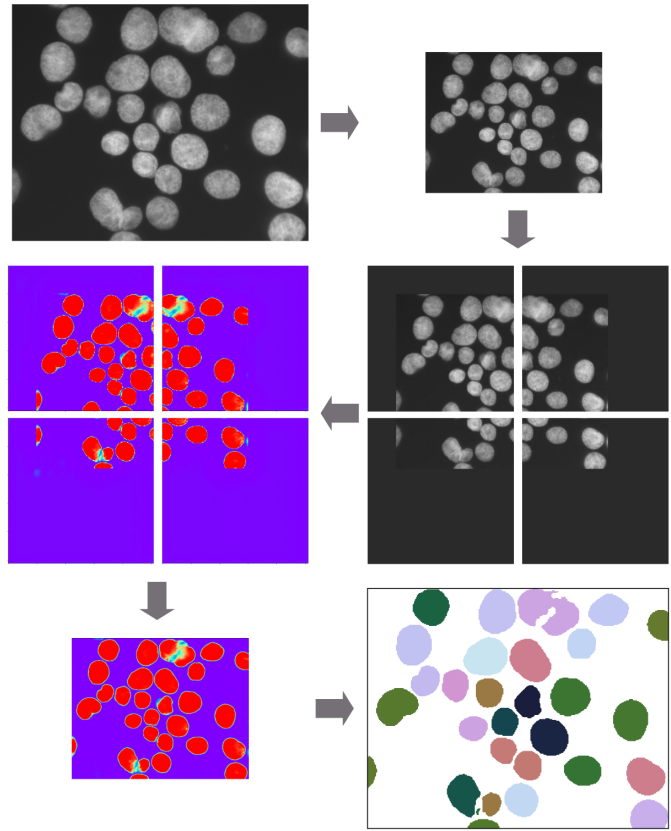

Fig. 2. Tiling strategy used to tile, segment and reassemble images to compare the segmentation performance of state-of-the-art deep learning architectures. First, all images are rescaled to have the same mean nuclei size across all images. Rescaled images are cropped into tiles using an overlap to avoid artifacts on the borders upon reconstruction. For each tile, segmentation prediction is inferred resulting in a probability map for each tile. After reassembling the predicted tiles, the predicted image is rescald, thresholded and labeled. Finally, small artifacts are removed based on a size threshold

| TYPE | REC | PREC | F1 | $\mu$DICE | AJI |
|---|---|---|---|---|---|
| Specimen: HaCaT | 0.99 | 0.84 | 0.91 | 0.99 | 0.82 |
| Specimen: Neuroblastoma | 0.95 | 0.78 | 0.85 | 0.97 | 0.63 |
| Specimen: Ganglioneuroma | 0.87 | 0.73 | 0.79 | 0.86 | 0.61 |
| Preparation: tissue section | 0.87 | 0.73 | 0.79 | 0.86 | 0.61 |
| Preparation: cell line cytospin | 0.99 | 0.85 | 0.92 | 0.99 | 0.87 |
| Preparation: grown cell line | 1 | 0.81 | 0.90 | 0.99 | 0.77 |
| Preparation: BM cytospin | 0.93 | 0.75 | 0.83 | 0.96 | 0.59 |
| Preparation: touch imprint | 0.99 | 0.84 | 0.91 | 0.97 | 0.74 |
| Challenge level: low | 0.99 | 0.88 | 0.93 | 0.99 | 0.88 |
| Challenge level: medium | 0.99 | 0.80 | 0.89 | 0.98 | 0.75 |
| Challenge level: high | 0.88 | 0.73 | 0.80 | 0.88 | 0.62 |
| **Diag./Prep./Challenge: All** | **0.92** | **0.77** | **0.84** | **0.92** | **0.74** |

TABLE I

COMPARISON BETWEEN SILVER-STANDARD AND GOLD-STANDARD ANNOTATIONS. GANGLIONEUROMA TISSUE CRYOSECTIONS AND NEUROBLASTOMA TUMOR TOUCH IMPRINTS, PRESENTING ARBITRARY NUCLEAR SHAPES, SHOW THE HIGHEST DIFFERENCE (F1 AND AJI SCORE). THIS IS ALSO EXPRESSED IN THE LOW MEDIUM- AND HIGH-LEVEL CHALLENGE F1/AJI SCORE WHEN COMPARED TO THE LOW-LEVEL FI/AJI SCORE.

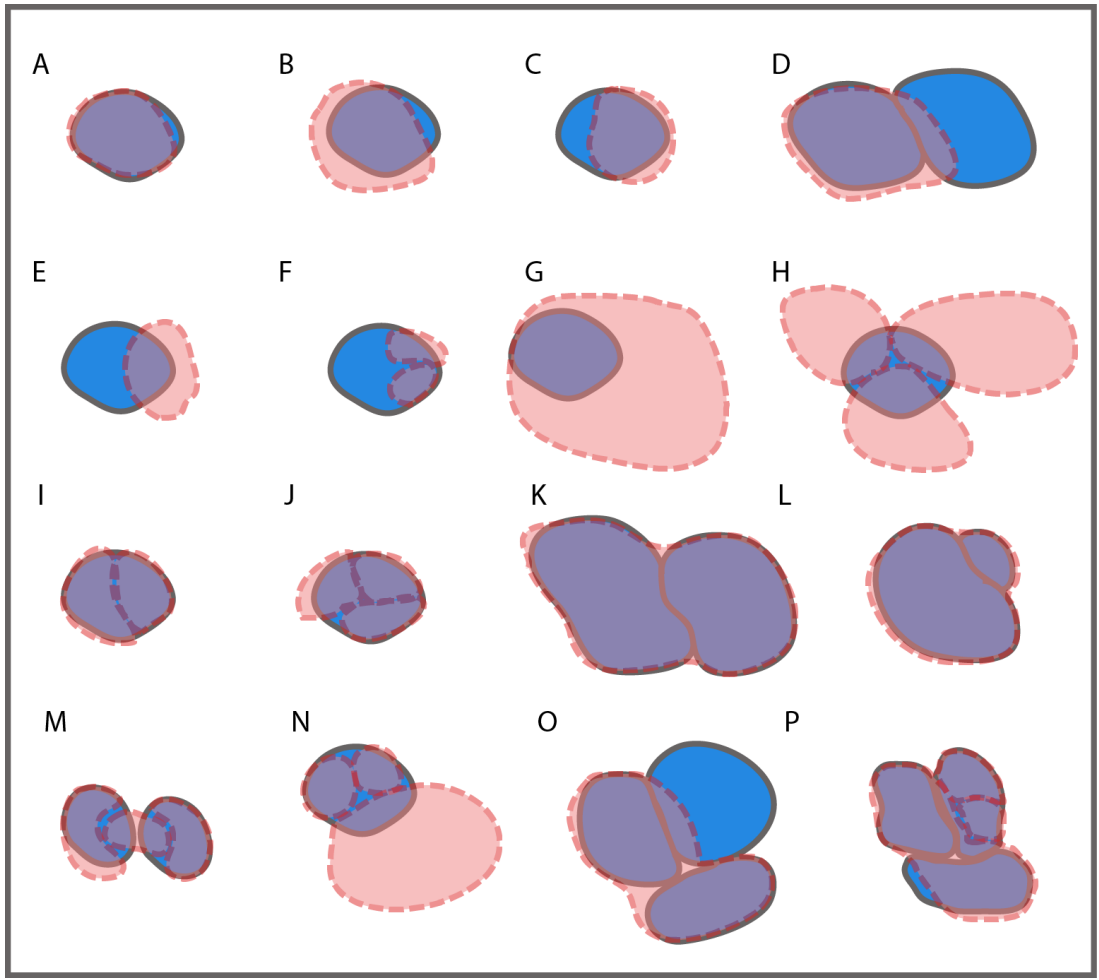

Fig. 3. Constructed ground truth and segmentation masks to illustrate cases with challenging evaluation. Ground truth objects are colored in blue, predicted objects in red with dotted edges. True positives (TP) and false negatives (FN) are always counted from objects of ground truth; False positives (FP) are always counted from objects of prediction. A-C: TP. D: One TP (left) and one FN (right). E: one FN and one FP. F-L: ground truth objects are FN and all predictions are FP. I-J: one over-segmentation (OS) K-L: two under-segmentations (US). M: one FN (left), one TP (right) and two FP (left two predictions). The right ground truth object is counted as TP because it is covered by exactly one prediction with a Jaccard Index (JI) of greater than 0.5, while the second prediction touching the ground truth object is not covered by the ground truth object for greater than 50%. The left object has a JI with the overlapping prediction of smaller than 0.5 and therefore accounts as FN. N: One FN and three FP. No OS since the two small predicted objects overlapped by the ground truth with more than 50% do not cover the ground truth object for more than 50%. O: three FN, one FP and two US. P: Three FN, three FP, two US and one OS.

Hyper-parameters were set based on experimental results and available resources.

| architecture | dataset | scaled | data augmentation | groundtruth | epochs | batchsize | input shape | learning rate | optimizer | additional parameters |
|---|---|---|---|---|---|---|---|---|---|---|
| U-Net | normal | yes | flip left/right, up/down | silver | 300 max | 2 | 1x256x256 | 0.001, adaptive | Adam | learning rate refinement strategy |
| U-Net | normal | yes | flip left/right, up/down | gold | 300 max | 2 | 1x256x256 | 0.001, adaptive | Adam | learning rate refinement strategy |
| U-Net | normal + artificial | yes | flip left/right, up/down | gold | 300 max | 2 | 1x256x256 | 0.001, adaptive | Adam | learning rate refinement strategy |
| U-Net | normal | no | flip left/right, up/down | gold | 300 max | 2 | 1x256x256 | 0.001, adaptive | Adam | learning rate refinement strategy |
| U-Net Resnet34 | normal | yes | - | silver | 300 | 2 | 3x256x256 | 0.001 | Adam | - |
| U-Net Resnet34 | normal | yes | - | gold | 300 | 2 | 3x256x256 | 0.001 | Adam | - |
| U-Net Resnet34 | normal + artificial | yes | - | gold | 100 | 2 | 3x256x256 | 0.001 | Adam | - |
| U-Net Resnet34 | normal | no | - | gold | 300 | 2 | 3x256x256 | 0.001 | Adam | - |
| Mask R-CNN | normal | yes | flip left/right, up/down | silver | 300 | 8 | 3x256x256 | 0.001 | SGD | - |
| Mask R-CNN | normal | yes | flip left/right, up/down | gold | 300 | 8 | 3x256x256 | 0.001 | SGD | - |
| Mask R-CNN | normal + artificial | yes | flip left/right, up/down | gold | 300 | 8 | 3x256x256 | 0.001 | SGD | - |
| Mask R-CNN | normal | no | flip left/right, up/down | gold | 300 | 8 | 3x256x256 | 0.001 | SGD | - |
| DeepCell | normal | yes | rotation, flip left/right, up/down | silver | 5 | 256 | 1x256x256 | 0.01 | SGD | window size 31x31 |
| DeepCell | normal | yes | rotation, flip left/right, up/down | gold | 25 | 256 | 1x256x256 | 0.01 | SGD | window size 61x61 |
| DeepCell | normal + artificial | yes | rotation, flip left/right, up/down | gold | 25 | 256 | 1x256x256 | 0.01 | SGD | window size 61x61 |
| DeepCell | normal | no | rotation, flip left/right, up/down | gold | 25 | 256 | 1x256x256 | 0.01 | SGD | - |
| Pix2pix | normal paired | yes | flip left/right, up/down, elastic deformation | gold | 130 | 1 | 256x512 | 0.0002 | Adam | - |

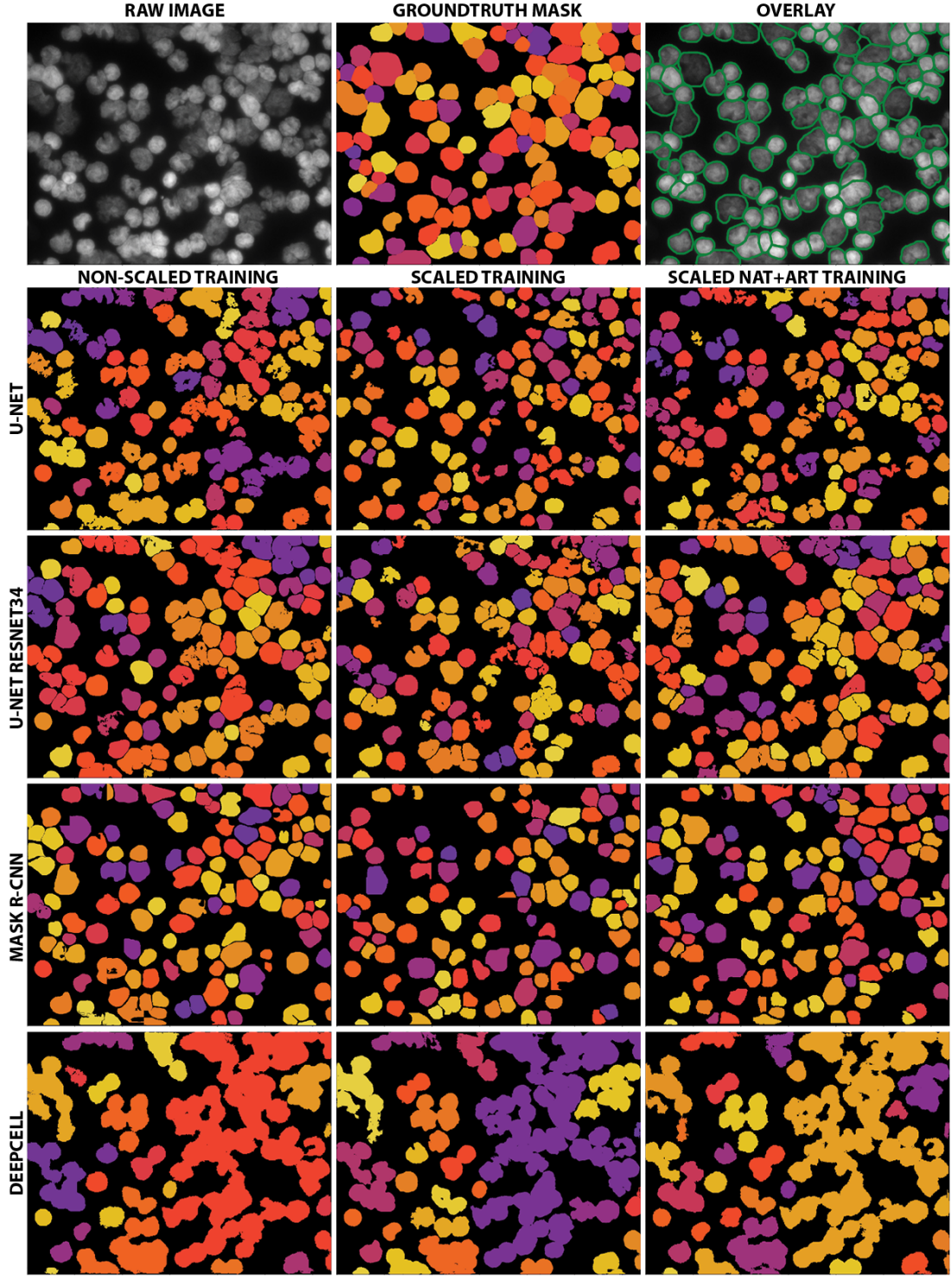

Fig. 4. Qualitative segmentation results on one bone marrow cytospin showing high nuclei density taken from the *gold-standard* test set. The first row depicts the raw image, the curated ground truth as well as the raw image overlayed with cell contours. Row two to five depict the instance segmentation results for all evaluated architectures. The columns represent the utilized training datasets: non-scaled natural, scaled natural and a combination of scaled natural and artificial images.